\documentclass{article} 
\usepackage{nips13submit_e,times}
\usepackage{hyperref, graphicx}
\usepackage{url}
\hypersetup{
    colorlinks=true,
    linkcolor=blue,
    filecolor=magenta,      
    urlcolor=cyan,
}

\title{Beating Atari with Natural Language\\Guided Reinforcement Learning}

\author{
Russell Kaplan\thanks{The authors contributed equally.}, Christopher Sauer\footnotemark[1], Alexander Sosa\footnotemark[1]\\
Department of Computer Science\\
Stanford University\\
Stanford, CA 94305 \\
\texttt{\{rjkaplan, cpsauer, aasosa\}@cs.stanford.edu} \\
}

\nipsfinalcopy 

\begin{document}

\maketitle

\begin{abstract}
We introduce the first deep reinforcement learning agent that learns to beat Atari games with the aid of natural language instructions. The agent uses a multimodal embedding between environment observations and natural language to self-monitor progress through a list of English instructions, granting itself reward for completing instructions in addition to increasing the game score. Our agent significantly outperforms Deep Q-Networks (DQNs), Asynchronous Advantage Actor-Critic (A3C) agents, and the best agents posted to OpenAI Gym \cite{gym} on what is often considered the hardest Atari 2600 environment \cite{intrinsic-motivation}:  \textsc{Montezuma's Revenge}.

\vspace{1em}

Videos of Trained \textsc{Montezuma's Revenge} Agents:

\href{https://www.youtube.com/embed/fqXNudaL1wY}{Our Best Current Model. Score 3500.}

\href{https://www.youtube.com/embed/NJMF4dfsWZw}{Best Model Currently on OpenAI Gym. Score 2500.}

\href{https://www.youtube.com/embed/lVHjQReKJNs}{Standard A3C Agent Fails to Learn. Score 0.}
\end{abstract}
\vspace{1.1em}

\begin{figure}[h]
\begin{center}
\includegraphics[width=4.7in]{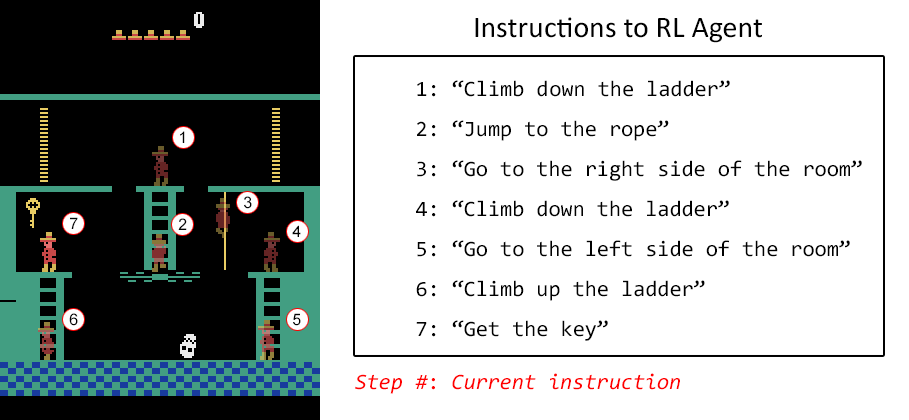}
\end{center}
\caption{Left: An agent exploring the first room of \textsc{Montezuma's Revenge}. Right: An example of the list of natural language instructions one might give the agent. The agent grants itself an additional reward after completing the current instruction. ``Completion" is learned by training a generalized multimodal embedding between game images and text.}
\end{figure}

\section{Introduction}

Humans do not typically learn to interact with the world in a vacuum, devoid of interaction with others, nor do we live in the stateless, single-example world of supervised learning. 

Instead, we live in a wonderfully complex and stateful world, where past actions influence present outcomes. In our learning, we benefit from the guidance of others, receiving arbitrarily high-level instruction in natural language--and learning to fill in the gaps between those instructions--as we navigate a world with varying sources of reward, both intrinsic and extrinsic. 

Building truly intelligent artificial agents will require that they be capable of acting in stateful worlds such as ours. They will also need to be capable of learning from and following instructions given by humans. Further, these instructions will need to be at whatever high-level specification is convenient---not just the low-level, completely specified instructions given in current programs.

Inspired by the dream of instructing artificial agents with the power of natural language, we set out to create an agent capable of learning from high-level English instruction as it learns to act in the stateful model-world of Atari games. 

We do so by combining techniques from natural language processing and deep reinforcement learning in two stages: In the first stage, the agent learns the meaning of English commands and how they map onto observations of game state. In the second stage, the agent explores the environment, progressing through the commands it has learned to understand and learning what actions are required to satisfy a given command. Intuitively, the first step corresponds to agreeing upon terms with the human providing instruction. The second step corresponds to learning to best fill in the implementation of those instructions.

\section{Background}

DeepMind shocked the deep and reinforcement learning communities in 2013 with the introduction of deep Q-learning. For the first time, reinforcement learning agents were learning from high-dimensional, visual input using convolutional neural networks \cite{DQNAtari}. With just screen pixels as input and the score as reward, their agents achieved superhuman performance on roughly half of Atari 2600 console games, most famously on Breakout.

\begin{figure}[h]
\begin{center}
\includegraphics[width=2.5in]{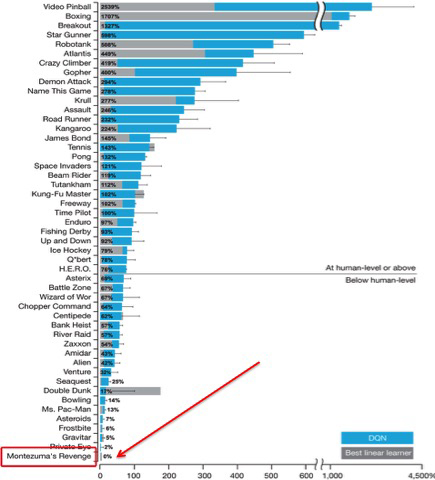}
\end{center}
\caption{Deep Q-Network performance on all Atari 2600 games, normalized against a human expert. The bottom-most game is \textsc{Montezuma's Revenge}. After playing it for 200 hours, DQN does no better than a random agent and scores no points. \cite{DQNNature}}
\end{figure}

The resulting models play impressively well; however, learning completely fails in games with extended times to rewards like \textsc{Montezuma's Revenge}, and the models require extensive exploration time to find good strategies, even for simpler games like Breakout. \cite{DQNNature,hierRL}.

Since then, reinforcement learning methods have been increasingly successful in a wide variety of tasks, particularly as a bridge to learn non-differentiable, stateful action. The range of applications extends far beyond the original paper, from early attention mechanisms to productive dialogue generation \cite{ShowAttendAndTell,RLDialogue}. There have also been significant improvements in the underlying learning architecture. In particular, there has been a shift away from the original deep Q-formulation toward the Asynchronous Advantage Actor-Critic because of improved learning speed and stability \cite{A3C}. The difference between Deep Q-Network (DQN) and Asynchronous Advantage Actor-Critic (A3C) will be discussed further in the next section.

Unfortunately, reinforcement learning agents still struggle to learn in environments with sparse rewards. \textsc{Montezuma's Revenge} has become the a key testing ground for potential improvements to the sparse reward problem and a very active area of research at DeepMind. Efforts to improve performance in the past few months have focused on adding additional rewards, e.g., curiosity, at various levels or adding additional model capabilities, e.g., memory \cite{NEC, FuN,AuxTasks,CountBased}. To the best of our knowledge, no one has previously tried guiding reinforcement learning with natural language instructions.

\subsection{Approaches to Reinforcement Learning}

Reinforcement learning is a broad conceptual framework that encapsulates what it means to learn to interact in a stateful, uncertain, and unknown world. In reinforcement learning, an agent observes some state from its environment at each time step and decides upon an action. The agent subsequently observes the updated state---affected both by the agent's actions and by external factors---and the agent may receive some reward. This cycle repeats until a termination condition is met. A successful reinforcement learning agent learns from  its experience in its environment to improve its acquisition of time-discounted reward in future runs.    

\begin{figure}[h]
\begin{center}
\includegraphics[width=2.5in]{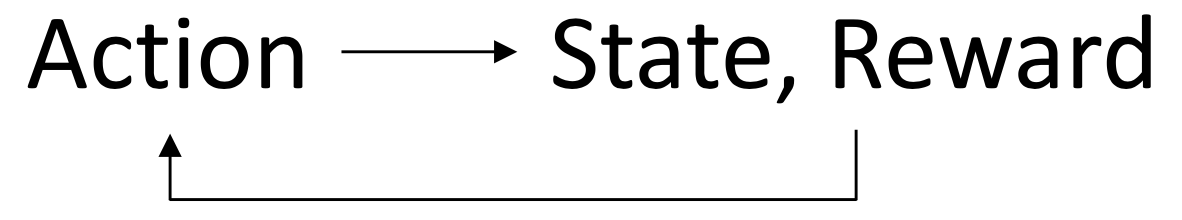}
\end{center}
\caption{Reinforcement learning cycle.}
\end{figure}

For all reinforcement learners there is the notion of the value of a state or action.  The objective is to maximize an exponentially time-discounted sum of future rewards. That is, there is some time-discounting factor $0  < \gamma \leq 1$, and the  value of a state or action---depending on the formulation---is the sum over future rewards of  $\gamma^{\Delta t}r_{\Delta t}$, where  ${\Delta t}$ is the time until the reward $r_{\Delta t}$ is earned. The time discounting combats value exploding to infinity over time and encourages faster pursuit of reward.

Despite this common framework, in practice there are two well-known but very different approaches for agent learning. We describe each below.  

\subsubsection{Deep Q-Learning: The Action-Value Formulation}

DeepMind's original Atari paper used an approach that they termed deep Q-learning, based on the older idea of Q-learning \cite{DQNAtari}. Q-learning agents learn a function $Q$, which takes the current state as input and applies a discounted value estimate for the rest of the game for each possible action.  At test time, the Q-learning agent simply picks the action with the highest estimated value for that state. At training time, it balances exploiting what it believes to be the best action with exploring other actions.

The function $Q$ is parameterized as a convolutional neural network and is trained to match the observed value of taking actions in a given state through standard backpropagation and gradient descent on a saved history of observed values. Of course, the value of a given action changes as the agent learns to play better in the future time steps, so this training is a moving target as the agent learns. 

\subsubsection{Policy Iteration and Asynchronous Advantage Actor-Critic: The Action-Distribution Formulation}

While Deep Q-Networks are famous from the original DeepMind Atari paper, policy-based approaches, in particular the Asynchronous Advantage Actor-Critic (A3C), now dominate most leaderboards on OpenAI Gym, a source of standardized reinforcement learning environments  \cite{gym}.

Instead of trying to assess action value, policy networks skip to directly learning policy, a function $\pi$ that maps a state to a distribution of actions. Policy networks maximize the expected, discounted reward of following that policy, which it can be shown equates to gradient descent with step size proportional to the discounted reward $R$ times the log of the probability $\pi$ assigned to the action. In contrast to Deep Q-Networks, learning can benefit from Thompson sampling or other techniques that leverage the degree of the network's uncertainty about the best next action. A popular variant, A3C, learns from many games in parallel, increases stability by subtracting an estimated state value $V(s)$ from  the reward multiplier $R$ in updates, and converges faster than other available options \cite{A3C,FuN}. 

\section{Approach and Experiments}

\subsection{Overview of Approach}

\begin{figure}[h]
\begin{center}
\includegraphics[width=5.5in]{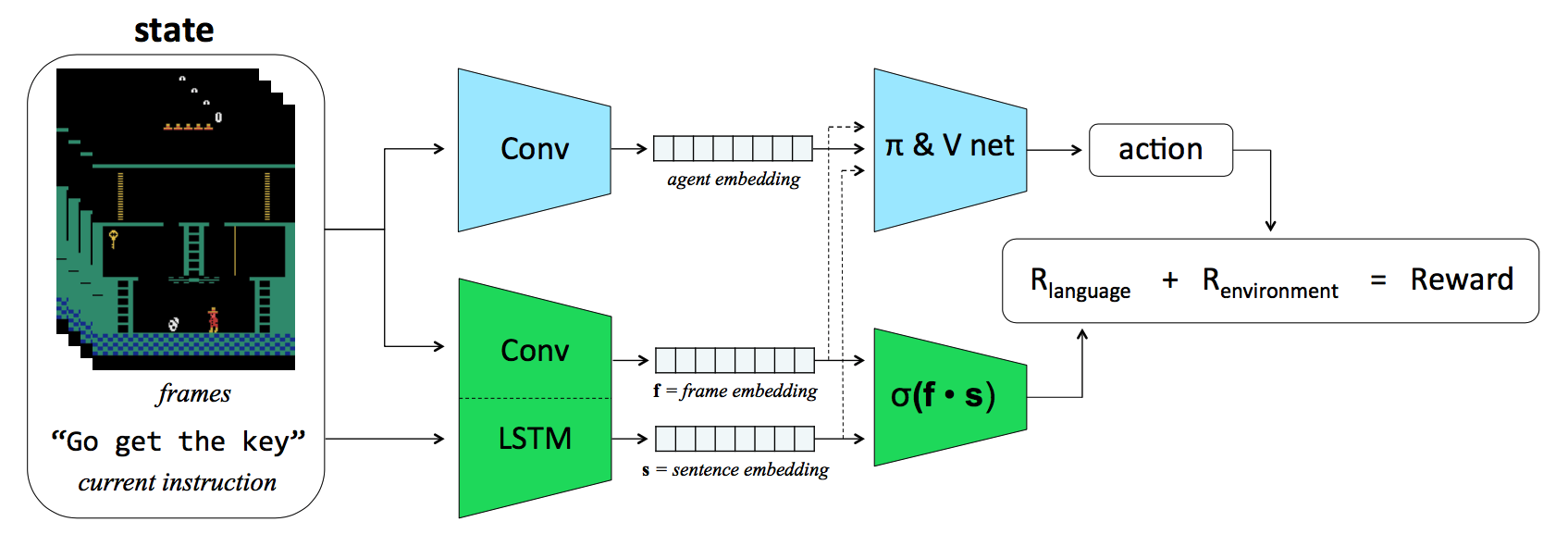}
\end{center}
\caption{The overall architecture of our natural language instructed agent at reinforcement learning time---described as the second step above. The agent's input state at a given frame is shown on the left, which consists of four recent frames---the last two frames and the 5th and 9th prior frames---and the current natural language instruction. As in a standard deep reinforcement learning agent, the state is run through a convolutional neural network and then fully connected policy and value networks---shown in blue---to produce an action and update. The multimodal embedding between frame pairs and instructions---trained in the first step above and shown in green---is used to determine if a natural language instruction has been satisfied by the past two frames. Satisfying an instruction moves the agent on to the next and leads to the agent giving itself a small additional reward. The frame and instruction sentence embedding are also passed as additional features to the network learning the policy and value. Intuitively, this equates to telling the agent (1) what is next expected of it, rather than leaving it to have to explore blindly for the next reward, and (2) how its progress is being measured against that command. Together these allow it to better generalize which actions are required to satisfy a given command.}
\end{figure}

\subsection{Baseline Models}

Being new to reinforcement learning, our first step was to train a robust baseline model on Atari Breakout. For this we used a standard Deep Q-Network based on an implementation in TensorFlow \cite{DQNTF,tensorflow2015-whitepaper}. However, after 30 hours of training, the agent failed to converge to one that could beat the game. Unable to reproduce DeepMind's results with DQN without weeks of training, we instead trained a stronger baseline using a model that is state-of-the-art for many reinforcement learning environments, including Breakout: Asynchronous Advantage Actor-Critic (A3C). We based the model on an implementation from Tensorpack \cite{Tensorpack} and relied on OpenAI's Gym framework for training reinforcement learning agents \cite{gym}. After training overnight, A3C successfully converged to a perfect Breakout score within 30 million frames of training.

\subsection{Viability of Sub-Task Rewards}
Equipped with this more promising baseline and caught up on our understanding of the state of the art, we set out to run a proof of concept on the usefulness of additional instruction to reinforcement learning agents by injecting an additional reward signal to the agent whenever in state that we determined was beneficial for its learning. Instead of having this additional reward based on completing a natural language instruction, we initially simply rewarded the agent for arriving at a game state where the ball's position was directly above the paddle's position. To accomplish this, we wrote a template-matching library for Breakout to return the positions of the ball and paddle within the environment. This turned out to be nontrivial given that the paddle in the Breakout environment gets smaller as the game progresses and can overlap with the walls and the ball in certain scenarios and that the ball changes colors based on y position, to name a few complications.  One develops a special appreciation for convolutional neural networks when forced to essentially hard-code one.  

With this additional template-matching reward for the agent obtaining the state of having the paddle under the ball, we saw a significant improvement in the rate of learning for early steps of the game. We saw even more improvement by also providing a negative reward whenever the agent lost a life (this reward signal is not provided by the environment directly). The additional reward signals for following our two hard-coded “instruction” helped the agent score more than 120 points after only 12,000 iterations (128 frames are played per iteration), more than three times better than ordinary A3C after the same amount of training time.

While these results for early training were promising, the additional reward signal nevertheless becomes less important as the agent progresses further through the environment. The additionally rewarded agent converges with the baseline A3C Breakout model after 36,000 iterations, and after 100,000 iterations it actually performs worse. We attribute this to both 1) the fact that simply keeping the paddle under the ball is not as useful of a strategy for later rounds when the ball moves more quickly than the paddle can, and 2) the fact that the baseline A3C agent learns so rapidly without additional reward signal because the reward is dense in the Breakout environment. The Breakout environment has such dense reward because every time a brick is broken, the agent is given a reward for the actions that led it to do so. Because the baseline A3C algorithm left little room for improvement for our additional reward signals, we decided to move on to the significantly harder \textsc{Montezuma's Revenge}.

\begin{figure}[h]
\begin{center}
\includegraphics[width=2.5in]{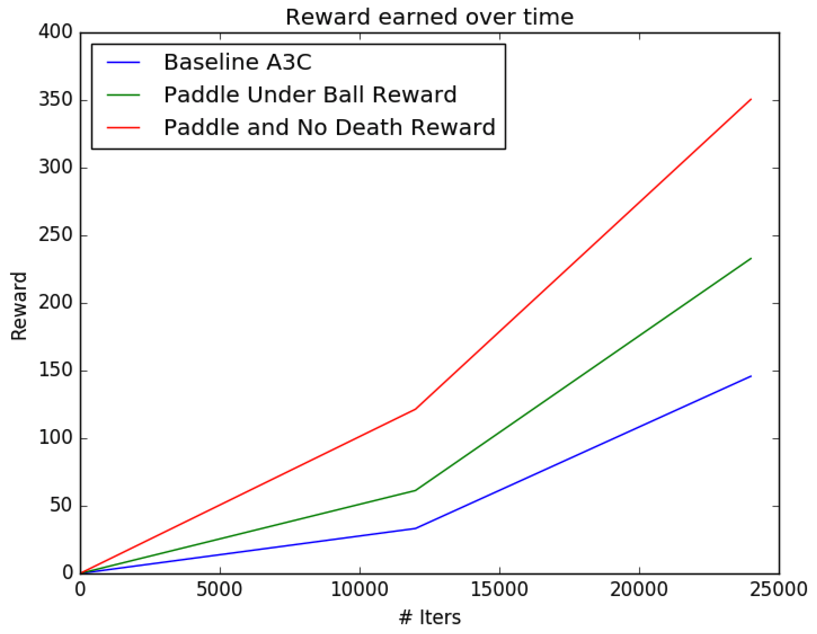}
\includegraphics[width=2.5in]{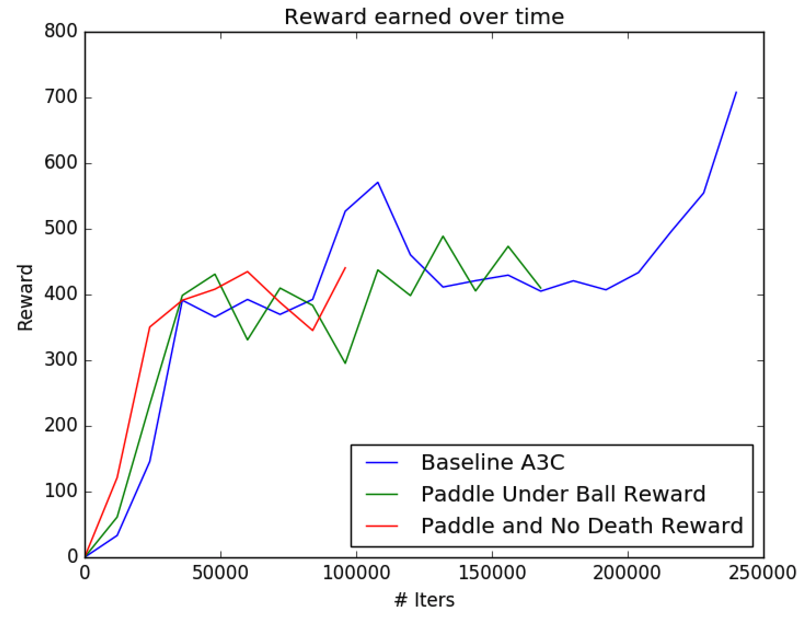}
\end{center}
\caption{Injecting additional reward in Breakout for keeping the paddle under the ball speeds initial learning greatly (left); however, once the agent masters the instructions, which describe only basic game play, the instructions cease to speed up learning (right). Having mastered Breakout, a more difficult environment---\textsc{Montezuma's Revenge}---is required for further reinforcement learning insight.}
\end{figure}

\subsection{Experiments in Multimodal Embeddings}

The aim of our initial experiments on Breakout was to use the simpler environment to refine and fully understand the sub-components of our natural language instructed model before we assembled them to tackle  \textsc{Montezuma's Revenge}.
Having experimented and debugged the reinforcement (A3C) and reward augmentation sub-components, the final component of our model we had not yet proven was the multimodal embedding mapping natural language descriptions and frames into a single embedding space where we could determine whether  the description applied to the frame.

We first generated a dataset of several  thousand frames by running  the A3C model we had just trained for Breakout  and using our template matching code to give relational descriptions of entities in the frame. For example, some of the possible frame descriptions included ``The ball is to the right of the paddle'' and ``The paddle is to the right of the ball.''

To correctly identify commands that are satisfied by a series of consecutive frames and to pass fixed size vectors describing those commands and frames to our learning agent, we need a multi-modal embedding between frames and sentences. The overall setup is shown in the green portion of Figure 4. Our network takes a pair of sequential frames and a sentence captured as a vector of words which may or may not describe that image. Frame pairs are taken to an embedding by a convolutional neural network. Instructions are taken to an embedding of the same size by a variety of techniques such as an LSTM, GRU, BiLSTM, BiGRU, and bag of words (BOW), described and compared in Figure 6.

\begin{figure}[h]
\begin{center}
\includegraphics[width=2.5in]{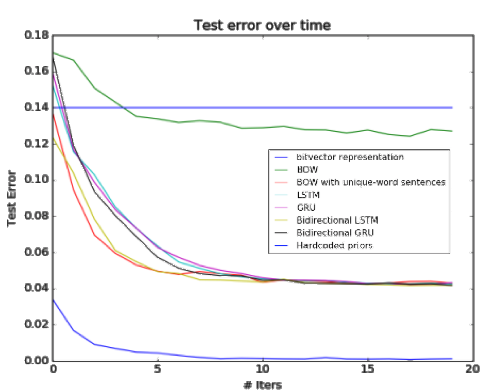}
\end{center}
\caption{Results from training the Breakout multimodal embedding network with different processing strategies for natural language input. The network is trained as a multi-label binary classifier that tries to answer, independently for each of the 23 Breakout natural language instructions our the dataset, ``Does this instruction correspond to these frames?" It predicts ``yes'' if the dot product between the frame embedding and that sentence's embedding is positive, and ``no'' if this dot product is negative. The straight line is what the test error would be if predictions are made based solely on the priors (i.e., ignoring the frames completely and just saying "yes" to each instruction if that instruction is labeled positively in more than half of the training examples). The bottom curve is what results from feeding the binary feature vector that was used to generate the natural language sentence in our dataset-generation stage directly into the multimodal network. We see that a simple bag-of-words approach barely does better than the priors. This is because in almost every sentence in the dataset word order matters: ``The ball is to the right of the paddle'' means something very different from ``The paddle is to the right of the ball,'' but the BOW representations are identical. LSTM, GRU, Bi-LSTM, Bi-GRU, and BOW with unique-word sentences (e.g., transforming ``paddle'' to ``paddle2'' whenever it was at the start of the sentence so that all sentences have unique words) all perform equally well, with roughly 4\% validation error.}
\end{figure}

The network aims to maximize the dot product between an embedding of the frames generated  and the sentence embeddings for all sentences that accurately describe a series of frames and to minimize the dot product with those sentence embeddings for sentences that do not apply to the frames. We choose a dot product between the frame and sentence embeddings over, for example, cosine distance because many sentences may match a single frame. We want the embedding to be able to measure whether an frame pair expresses one meaning component---the one given by the sentence vector---without penalizing its score for expressing another meaning component, as would occur with the normalization terms in cosine similarity. 

\subsection{\textsc{Montezuma's Revenge}}

Unlike Breakout, the \textsc{Montezuma's Revenge} environment has very sparse reward signal. Refer to Figure 1 to get an idea: Even after the seven listed instructions are followed, the agent still has zero reward. Only after reaching the key is any reward received, and walking off any of the intermediary platforms or touching the skull on the way to the key results in death. It is no surprise that given these challenges, DQN and A3C both struggle to learn anything.

We ran the baseline A3C algorithm on \textsc{Montezuma's Revenge} and found that after 72 hours of gameplay and more than 200,000,000 observed frames, the agent achieved a consistent score of 0 in the environment. The actions the trained agent would choose tended to result in near-immediate death. With this much less impressive baseline model, we repeated the experiment of matched rewards that we ran on the Breakout environment. We wrote a new template-matching library for \textsc{Montezuma's Revenge} and formulated a new series of commands that we devised to be not only useful to the agent to escape the first room but also generalizable across different rooms present in the level. (See Appendix A for the full list.) Like Breakout, we started with an easier experiment: Rather than learn to use these natural language commands directly, we first rewarded the agent for simply reaching different manually specified locations of the room shown in Figure 1. With this approach, we were able to achieve a mean score of 400 before exhausting our instructions and consistently have the agent make it out of the first room. These results were promising to our eventual end-goal since they confirmed that some source of informative auxiliary reward helps the agent reach farther.

\subsection{Dataset Generation}
Having shown that subtask rewards were very promising for an reinforcement learning agent learning to play \textsc{Montezuma's Revenge}, we were next tasked with generating a dataset to learn mappings between natural language descriptions of state and raw pixels from the environment. No such dataset existed, however, so we created our own mappings of game state to natural language descriptions that they satisfied. To do so, we played fully through the game several different times, saving frames of game state for each run-through. Utilizing the template-matching code, we could generate lists of chosen commands that were satisfied by a given series of consecutive frames. After several playthroughs, we amassed 15,000 training frames and held out another 3,000 frames---a full separate playthrough---for validation.

\subsection{Learning Frame-Command Mappings with Bimodal Network}

As with Breakout, the first phase of training constituted creating a multimodal embedding between frame pairs depicting the player's motion and command statements in Appendix A. The embedding was trained such that when commands were satisfied by the frames, there was positive dot product between the frame and command embeddings. When the commands were not satisfied, the dot product was trained to be negative. We used a LSTM with word-level embeddings to extract command embeddings and a convolutional neural network running over pairs of frames stacked on the channel dimension for the frame embeddings. 

\subsubsection{Evidence of Generalization}

Given the complexity of the bimodal embedding model and its access to a training dataset containing all the rooms for the first level, we wanted to demonstrate that the bimodal was actually learning to generalize the meaning of the commands, not just requiring that the agent be in the exact same position as in the training data to classify a command as complete.

Of course, by checking against the unseen playthrough as validation---which contains many frames that differ from those in the training set---we were already testing for generalization within a room. However, we hoped that the command-frame embeddings might be able to generalize across to unseen rooms. To test this, we retrained our bimodal embeddings, this time removing access to all training data for the second room, but still including frames from the second room in the validation set. The game elements seen in the second room (e.g., ladders) are still present individually in other rooms, so this tests whether the embedding can understand them in their new configuration when tested on the unseen second room.

\begin{table}[ht]
\centering 
\vspace{.5em}
\begin{tabular}{c c c c} 

Validation Error Type & Full Training Set & Training Set Excluding Room 2 \\ [0.5ex] 
\hline \\ [-1.5ex]
Overall validation error percentage & 0.0019713 & 0.0012545 \\
Climb down the ladder & 0 & 0 \\
Climb up the ladder & 0.066667 & 0.066667 \\
Get the coin & 0 & 0 \\
Get the key & 0 & 0 \\
Get the sword & 0 & 0 \\
Get the torch & 0 & 0 \\
Go between the lasers & 0 & 0 \\
Go to the bottom of the room & 0 & 0 \\
Go to the bottom room & 0 & 0 \\
Go to the center of the room & 0 & 0 \\
Go to the left room & 0 & 0 \\
Go to the left side of the room & 0.066667 & 0 \\
Go to the right room & 0 & 0 \\
Go to the right side of the room & 0 & 0 \\
Go to the top of the room & 0 & 0.066667 \\
Go to the top room & 0 & 0 \\
Jump to the rope & 0 & 0 \\
Use the key & 0 & 0 \\
\hline 
\end{tabular}
\label{table:nonlin} 
\caption{Multimodal embedding error rates after training for 100 epochs. Note that holding out room 2 in the training set has little effect on the embedding accuracy, even through the validation set contains room 2. }
\end{table}

The embedding scheme seems to maintain its extremely low error rate even when tested on a room it did not observe during training time. This provides some evidence that the embedding generalizes to some degree across rooms as opposed to just behavior within a room.

It is neat that the embeddings appear to learn to have captured common meaning across rooms, and further, that a relatively simple set of commands gives us as human instructors enough expressiveness to write instructions through the whole first level. Furthermore, the agent can then learn to use this common representation as a basis for generalizing its actions across rooms given a command. For example, if told to go to the right room, the agent might apply the knowledge gained in first figuring out that instruction's meaning to choose the correct actions on its second appearance in a different room. Indeed, we see this behavior qualitatively as we watch the agent train. Climbing down the first ladder takes considerable effort, since the agent needs to discover it must hold the same arrow key for the duration. On subsequent encounters, the agent often completes the command on the first try; however, learning to go up a ladder for the first time takes longer to learn.

\subsection{Run-Time Learning From Natural Language Reward Using Multimodal Embedding}

With training of the multimodal embedding complete, we then move to the reinforcement learning stage. For each A3C worker, we load the embedding weights and a list of commands for the agent to sequentially complete. We calculate whether a command was completed by using our pretrained bimodal embeddings, passing the current observed state and the current command to accomplish through the network and marking the command as completed if the dot product between the resulting frame and command embeddings is positive. If we note that a command has completed, we give our reinforcement learning agent an additional reward for successfully completing the task and continue on to the next command, also feeding the embeddings into the learning agent as additional features, as described in Figure 4.

\subsection{Final Results}

\begin{table}[ht]
\centering 
\vspace{.5em}
\begin{tabular}{c c c c} 

Algorithm & Score (environment reward) \\ [0.5ex] 
\hline \\ [-1.5ex]
Nature DQN & 0.0 \\
A3C & 0.1 \\
MFEC & 76.4 \\
NEC & 42.1 \\
Prioritised Replay & 0.0 \\
Q*(lambda) & 0.4 \\
Retrace(lambda) & 2.6 \\
\textbf{Instructed Reinforcement Learner (ours)} & \textbf{500.0} \\ [1ex] 
\hline 
\end{tabular}
\label{table:nonlin} 
\caption{Results on \textsc{Montezuma's Revenge} after training for 10 million frames. All scores other than ours are pulled from \cite{NEC}, and all of the reported scores in \cite{NEC} are included here. These are all the reported scores we could find for \textsc{Montezuma's Revenge} agents trained for 10 million frames. }
\end{table}

\begin{table}[ht]
\centering 
\vspace{.5em}
\begin{tabular}{c c c c} 

Algorithm & Score (environment reward) \\ [0.5ex] 
\hline \\ [-1.5ex]
Itsukara's algorithm, \#2 on Gym leaderboard & 1284.0\\
Pkumusic's algorithm, \#1 on Gym leaderboard & 2500.0\\
\textbf{Instructed Reinforcement Learner (ours)} & \textbf{3500.0} \\ [1ex] 
\hline 
\end{tabular}
\label{table:nonlin} 
\caption{Comparison of results on \textsc{Montezuma's Revenge} between our agent and the OpenAI Gym leaderboard, with no limit on the number of frames trained on. Gym's leaderboard does not specify the number of frames agents were trained on. Our agent in this table learned for 60 million frames.}
\end{table}

Table 2 shows a comparison of different reinforcement learning algorithms and ours after 10 million frames of training. Table 3 compares our results to the leaderboard on OpenAI's Gym. Our agent is the clear and compelling winner of both comparisons.

It is important to note that these tables do not represent an apples-to-apples comparison. Our agent's ``environment'' for \textsc{Montezuma's Revenge} is different from that of all others: We include a helpful natural language instruction in addition to the visible frames. Clearly, this makes learning easier than with no language guidance, but that is entirely the point. We think our results are significant for two reasons. First, the type of supervision our agent received from the language has shown to generalize to new states---the agent is able to use language to assist navigation through frames that it has never seen before and that were not included in the training set of the multimodal embedding, so it is not merely memorizing. Second, the type of language supervision given to our agent is exactly the type of supervision that is reasonable to expect an agent would be able to receive in the real world.

There are other, orthogonal ways of adding extra reward to the training process that achieve strong results on \textsc{Montezuma's Revenge} as well. In \cite{intrinsic-motivation}, which uses Intrinsic Motivation as a source of bonus reward, the authors report that their best run achieved a score of 6600 after 100 million training frames. (This remains the highest reported score for a reinforcement learning agent on \textsc{Montezuma's Revenge}.)  It is important to note that the Intrinsic Motivation agent explores up to the same depth of rooms as ours when each is trained for 10 million training frames. (Unfortunately, they do not report the score after 10 million frames of training, just the rooms explored.) We were not able to train for 100 million frames to provide a more direct comparison with their final results, due to limited computational budget. 

\section{Conclusion}

We present a novel framework for training reinforcement learning agents that allows the agent to learn from instruction in natural language. It is a promising start to cooperation between reinforcement learning agents and their human trainers: The agent achieves impressive scores in relatively few frames where traditional agents fail. We think this approach will be even more fruitful when applied to reinforcement learning in the real world, for example, in robotics. This is because extremely rich, labeled datasets already exist for real-world images, allowing a much more sophisticated multimodal embedding between image and language to be learned than what is achievable with our synthetic dataset. In addition, many tasks in robotics also suffer from the sparse-reward problem, which our approach is specifically designed to address. We imagine it would be quite useful to have an intelligent robot that can be instructed by any human, not just an expert programmer, to quickly learn new tasks.

We hope to explore these ideas in future work. We also think it would be interesting to combine our additional reward mechanism with other sources of auxiliary reward, like Intrinsic Motivation, which could quite feasibly achieve state-of-the-art results on many challenging environments due to their complementary and orthogonal nature.

{\small
\bibliographystyle{ieee}
\bibliography{main}
}

\section*{Appendix A: Full List of Commands Usable for \textsc{Montezuma's Revenge}}

\begin{verbatim}
Climb down the ladder
Climb up the ladder
Get the key
Use the key
Get the sword
Get the torch
Get the coin
Jump to the rope
Go to the left side of the room
Go to the right side of the room
Go to the bottom of the room
Go to the top of the room
Go to the center of the room
Go to the bottom room
Go to the left room
Go to the right room
Go to the top room
Go between the lasers
\end{verbatim}

Final architectures for \textsc{Montezuma's Revenge}:\\

Multimodal embedding network:
\begin{verbatim}
Frame head:
Conv[5x5, 32 filters]
ReLU
MaxPool[2x2]
Conv[5x5, 32 filters]
ReLU
MaxPool[2x2]
Conv[4x4, 64 filters]
ReLU
MaxPool[2x2]
Conv[3x3, 64 filters]
FullyConnected[Output dimension 10]
PReLU
FullyConnected[Output dimension 10]

Sentence head:
Word vectors of size 12 ->
LSTM with hidden state size 10
\end{verbatim}

RL policy and value network:
\begin{verbatim}
Conv[5x5, 32 filters]
ReLU
MaxPool[2x2]
Conv[5x5, 32 filters]
ReLU
MaxPool[2x2]
Conv[4x4, 64 filters]
ReLU
MaxPool[2x2]
Conv[3x3, 64 filters]
FullyConnected[Output dimension 10]
PReLU
-> (Policy) FullyConnected[Output dimension 10]
-> (Value) FullyConnected[Output dimension 1]
\end{verbatim}

Full list of instructions given to the agent.
\begin{verbatim}
## Level 1
# Room 1

Climb down the ladder
Jump to the rope
Go to the right side of the room
Climb down the ladder
Go to the bottom of the room
Go to the center of the room
Go to the left side of the room
Climb up the ladder
Get the key
Climb down the ladder
Go to the bottom of the room
Go to the center of the room
Go to the right side of the room
Climb up the ladder
Jump to the rope
Go to the center of the room
Climb up the ladder
Go to the top of the room
Go to the right side of the room
Use the key
Go to the right room

# Room 2
Go to the center of the room
Climb down the ladder
Go to the bottom of the room
Go to the bottom room

# Room 3
Go to the left side of the room
Get the sword
Go to the center of the room
Go to the right side of the room
Go to the right room

# Room 4
Go between the lasers
Go to the center of the room
Get the key
Go between the lasers
Go to the center of the room
Climb down the ladder
Go to the bottom of the room
Go to the bottom room

# Room 5
Go to the left side of the room
Go to the left room

# Room 6
Go between the lasers
Go to the center of the room
Go between the lasers
Go to the left side of the room
Go to the left room

# Room 7
Go to the center of the room
Climb up the ladder
Go to the top room

# Room 8 (torch room) — 709
Use the key
Jump to the rope
Go to the center of the room
Jump to the rope
Go to the top of the room
Go to the center of the room
Get the torch
Jump to the rope
Go to the center of the room
Jump to the rope
Go to the bottom of the room
Go to the bottom room
\end{verbatim}

\end{document}